\documentclass[10pt,a4paper,twocolumn]{article}

\usepackage[T1]{fontenc}
\usepackage[utf8]{inputenc}
\usepackage{newtxtext,newtxmath}
\usepackage[top=2.2cm,bottom=2.2cm,left=1.8cm,right=1.8cm,columnsep=0.7cm,headsep=0.5cm]{geometry}
\usepackage{setspace}
\usepackage{microtype}
\usepackage{graphicx}
\usepackage{booktabs}
\usepackage{threeparttable}
\usepackage{caption}
\usepackage{amsmath}
\usepackage{fancyhdr}
\usepackage{titlesec}
\usepackage{xurl}
\usepackage[hidelinks]{hyperref}
\usepackage[natbibapa]{apacite}

\titleformat{\section}{\normalsize\bfseries\raggedright\MakeUppercase}{\thesection}{0.6em}{}
\titleformat{\subsection}{\normalsize\bfseries\raggedright}{\thesubsection}{0.6em}{}
\titleformat{\subsubsection}{\normalsize\itshape\raggedright}{\thesubsubsection}{0.6em}{}
\titlespacing*{\section}{0pt}{12pt}{5pt}
\titlespacing*{\subsection}{0pt}{9pt}{3pt}
\titlespacing*{\subsubsection}{0pt}{7pt}{2pt}
\titlespacing*{\paragraph}{0pt}{4pt}{0.6em}

\fancypagestyle{plain}{\fancyhf{}\fancyfoot[C]{\footnotesize\thepage}}

\hypersetup{pdftitle={From Exposure to Expectation: Frequency, Surprisal, and Language Across Development in Spanish},
            pdfauthor={Francisco Portillo L\'opez}}

\begin{document}
\shortcites{bates2015,canete2020,cuetos2011,delarosa2022,duchon2013,frank2017,kuznetsova2017,ramscar2013,ramscar2010,shliazhko2022,siegelman2022,siegelman2025,warstadt2023,wilcox2020,wilcox2023,xu2023,goodman2008,hills2010,vanheuven2014}

\twocolumn[
\begin{@twocolumnfalse}
\begin{center}
{\LARGE\bfseries From Exposure to Expectation: Frequency, Surprisal,\\[2pt] and Language Across Development in Spanish\par}
\vspace{14pt}
{\large Francisco Portillo L\'opez\par}
\vspace{4pt}
{\small Universidad de Navarra, Pamplona, Spain\\
\href{mailto:fportillolo@alumni.unav.es}{fportillolo@alumni.unav.es} \quad ORCID: \href{https://orcid.org/0009-0001-2580-5177}{0009-0001-2580-5177}\par}
\end{center}

\vspace{6pt}
\begin{center}\begin{minipage}{0.88\textwidth}
\begin{center}\textbf{Abstract}\end{center}
\vspace{-4pt}
{\small
\noindent Surprisal, the negative log-probability that a language model assigns to a word given its context, reliably predicts how long adults take to read that word. Does it also tell us when children learn it? Frequency and surprisal are often treated as two faces of linguistic experience, but they measure different things: frequency reflects how much a learner has heard a word, surprisal how predictable a particular occurrence is. We compared the two in two corpus studies of Spanish. In Study~1, we modeled the age of acquisition (AoA) of 225 nouns from their frequency in child-directed speech and from surprisal computed with three language models (mGPT, BETO, BERTIN), both in a carrier sentence and in naturalistic child-directed utterances. Frequency strongly predicted AoA ($\rho = -.59$). All six surprisal measures were also correlated with AoA ($\rho = .14$ to $.31$), but this association was largely shared with frequency: beyond frequency and word length, none explained reliable additional variance ($\Delta R^2 \le .011$; all Holm-corrected $p > .29$). In Study~2, we modeled total fixation durations of 45 Chilean Spanish readers in the MECO Wave~2 eye-tracking corpus. mGPT surprisal predicted longer fixations beyond frequency and length with two independent frequency norms and after controlling for spillover. The effect weakened, but remained reliable, when the model was given only the previous three words of context, so it does not depend on the long contexts available in reading. Taken together, the results suggest that cumulative exposure is what best predicts when early words are learned, while contextual predictability shapes how words are processed once they are known. They also suggest that a language model's fit to adult reading is not, on its own, evidence that it captures what matters for acquisition.\par}

\vspace{6pt}

\vspace{6pt}
\noindent{\footnotesize\textit{Note on this version.} This version corrects errors in the computation of several surprisal measures in the first arXiv version (v1), most importantly a tokenization error in Study~2. All analyses were recomputed; some estimates and one conclusion changed. The errors and their consequences are documented in \texttt{CORRECTIONS.md} in the project repository (see Data and Code Availability).}
\end{minipage}\end{center}
\vspace{14pt}
\end{@twocolumnfalse}
]
\thispagestyle{plain}

\section{Introduction}

Why do children learn some words before others? And why is a word that we already know easy to read in one sentence and harder in the next? The two questions are usually studied by separate literatures, with different populations, corpora and dependent measures. Yet both are about how the statistics of linguistic experience shape behavior, and both are now often approached with the same computational tool: surprisal estimated by a language model.

Surprisal is the negative logarithm of the probability of a word given its context. In information-theoretic accounts of comprehension, a word with low conditional probability carries more information and should take more effort to process \citep{hale2001,levy2008}. Surprisal does predict reading times, roughly linearly in log probability \citep{smith2013,wilcox2020}, and this holds across many languages \citep{devarda2023,wilcox2023,xu2023}, including Spanish \citep{devarda2022,wilcox2023}. These results have made model-derived surprisal a standard way of asking whether what a language model learns corresponds to anything psychologically real.

Success in adult reading, however, does not mean that surprisal plays the same role in acquisition. Age of acquisition (AoA) is not a measure of how hard a word is to process on one occasion; it is a developmental outcome, the point at which a word becomes part of a child's vocabulary after many encounters with it. This difference in timescale matters. Frequency measures how much experience a learner has had with a word. Surprisal measures how predictable one occurrence is in its context. The two are related, because frequent words also tend to be predictable, but they are not the same thing: a frequent word can still be unexpected in a particular sentence, and a rare word can be almost certain in the right one.

The distinction is central to usage-based accounts of lexical development. On these accounts, each encounter with a form strengthens, or entrenches, its representation \citep{bybee2010,langacker1987,tomasello2003}. Cumulative frequency should then be a natural predictor of acquisition order, simply because children have more opportunities to build representations of words they hear often. Contextual predictability may still matter, but there is no reason to expect it to be the property most closely tied to when a word is learned.

Error-driven, or discriminative, theories of learning make a different prediction \citep{ramscar2007,ramscar2010,ramscar2013}. If what a learner takes from an encounter depends on how far it departs from their expectations, then encounters in which a word is poorly predicted should be especially informative. On the simplest version of this view, words that tend to occur in less predictable contexts should be learned faster, so surprisal should be related to AoA (negatively) even after frequency is controlled. A null relation between surprisal and AoA is therefore not what every learning theory predicts. The present studies were not designed to decide between these accounts. Their aim is more basic: to establish what the relation between model surprisal and AoA actually looks like in Spanish once frequency is taken into account, which is something both accounts must eventually be consistent with.

Expectation-based theories of processing deal with a later stage. An adult reader already has a large lexicon and grammar and uses context to anticipate what comes next \citep{hale2001,levy2008}. For such a reader, surprisal is a direct measure of the mismatch between expectation and input, and its relation to reading time is naturally read as evidence that predictability affects the effort needed to process a word that is already known. Children learning their first words, in contrast, are still building the system that will later support those predictions. A model trained mostly on adult text may therefore approximate a mature reader's expectations without capturing the experience that matters most for early word learning. Work on child-directed speech has consistently linked frequency and related distributional measures to the timing of word learning \citep{goodman2008,hills2010}, while model surprisal has been very successful with adult reading \citep{smith2013,wilcox2020}. What remains unclear is whether these findings reflect one statistical mechanism or different aspects of experience that matter at different points in development.

\subsection{The present studies}

We address this question in Spanish with two corpus studies. Study~1 asks whether model surprisal predicts the AoA of 225 nouns beyond their frequency and contextual diversity in child-directed speech (CDS). We estimated surprisal with three language models that differ in architecture and training data (mGPT, BETO and BERTIN), so that the result does not hinge on one model, and in two kinds of context: a fixed carrier sentence and naturalistic CDS utterances. Evaluating masked models alongside an autoregressive one also lets us ask whether any relation is specific to left-to-right prediction; architecture and model size are known to affect how well surprisal fits human reading \citep{oh2023}. Because all three models work with subword units, we computed word-level surprisal by summing over a word's subword tokens, the standard approach; subword tokenization does not seem to put surprisal at a systematic disadvantage as a predictor of reading \citep{nair2023}, but it is a modeling choice that we report explicitly.

Study~2 asks whether surprisal from one of the same models, mGPT, predicts adult reading in Spanish. Using the Chilean Spanish subsample of the Multilingual Eye-movement Corpus (MECO Wave~2), we test whether surprisal explains variation in fixation duration beyond word frequency and length, with two independent frequency norms. Study~2 also addresses an obvious asymmetry between the two settings. In reading, the model can use the whole preceding passage, whereas a carrier sentence gives it only a few words. If surprisal predicts reading but not acquisition only because it has more context to work with in reading, the reading effect should disappear when the model's context is cut down. We therefore recomputed reading-time surprisal with the context limited to the last 20, 10, 5 or 3 words.

Our question is not whether surprisal is ``important'' for language in general. It is whether the same computational estimate of predictability has the same explanatory value for two different processes: the establishment of lexical knowledge during development, and the processing of words in a system where that knowledge is already in place. If cumulative exposure is what matters for early word learning, frequency should explain much of the variation in AoA and surprisal should add little once frequency is controlled. If contextual prediction matters for mature processing, mGPT surprisal should predict reading difficulty beyond frequency and length, and it should do so even with short contexts.

\section{Study 1: Surprisal and Age of Acquisition}

\subsection{Method}

\subsubsection{Nouns and age of acquisition}

AoA estimates came from the Spanish CDI norms in Wordbank \citep{frank2017}. Of the 288 noun items, we kept the 249 that consist of a single word (excluding items such as \textit{dedo/s} `finger/s' or \textit{coche polic\'ia} `police car'), and of these the 225 that occur at least once in the corpus described below. AoA ranged from 16 to 30 months ($M = 24.7$, $SD = 2.7$) and served as the dependent measure throughout.

\subsubsection{Corpus and distributional measures}

We used all utterances addressed to the target child by mothers, fathers and other adults in four Spanish corpora in CHILDES \citep{macwhinney2000}: Aguirre, Linaza, Ornat and Vila. Together they contain 34,930 utterances (135,583 word tokens) from 204 transcripts of 7 children aged 11 to 59 months. \textit{Frequency} was the number of times the exact word form occurred in this corpus (median 10, range 1--201). \textit{Contextual diversity} was the number of transcripts in which the word occurred (median 5, range 1--64). Following earlier work on CDS \citep{hills2010}, contextual diversity is meant to capture how widely a word is spread across different situations, as opposed to how often it is repeated. Because the two measures were highly correlated, we examined each on its own and controlling for the other.

\subsubsection{Language models and surprisal}

We used three pretrained models without fine-tuning: mGPT, a multilingual autoregressive model trained on 60 languages \citep{shliazhko2022}; BETO, a BERT model trained on a large Spanish corpus \citep{canete2020}; and BERTIN, a RoBERTa model trained on the Spanish portion of mC4 \citep{delarosa2022}. For mGPT, surprisal was the negative log probability (in bits) of the target word given the words before it. The context and the target were tokenized as in running text, so that the target kept its leading space, and the surprisal of the word was the sum over its subword tokens. For BETO and BERTIN, which predict masked tokens from both sides, we used the pseudo-log-likelihood of the target with within-word left-to-right masking \citep[PLL-word-l2r;][]{kauf2023}: for each subword token of the target, that token and the later tokens of the same word were masked, the token was predicted from the rest of the utterance, and the resulting surprisals were summed.

We estimated surprisal in two kinds of context. In the \textit{frame} condition, each noun completed the sentence \textit{la mam\'a cogi\'o} [\textit{el/la/los/las}] \_\_\_ (`mom picked up the \_\_\_'), with the article that agrees with the noun. Because the noun is the last word of the frame, all three models see only the four words before it. In the \textit{natural} condition, we sampled up to ten CDS utterances per noun in which the noun occurs as a whole word but not as the first word, so that there was always some preceding context. This gave 1,574 utterances for 224 nouns (one noun occurred only utterance-initially); 154 nouns had at least five utterances and 111 had ten. Surprisal was averaged over the utterances of each noun. The same utterances were used for the three models.

\subsubsection{Statistical analysis}

We first computed Spearman correlations between AoA and each predictor. To ask whether surprisal adds to exposure, we then fit linear regressions of AoA on log frequency, $\log(1 + \text{count})$, and word length in letters, added each surprisal measure in a second step, and report the change in $R^2$ and the standardized coefficient of surprisal; all variables were $z$-scored. Since there were six surprisal measures (three models $\times$ two kinds of context), we report Holm-corrected $p$ values alongside the uncorrected ones \citep{holm1979}. Two robustness checks followed: rank-based (Spearman) partial correlations between surprisal and AoA, controlling for log frequency and length, which are less sensitive to the skewed distributions of some surprisal measures; and the natural-context analyses restricted to the 154 nouns with at least five utterances.

\subsection{Results}

\subsubsection{Frequency and contextual diversity}

\begin{figure}[t]
\centering
\includegraphics[width=\columnwidth]{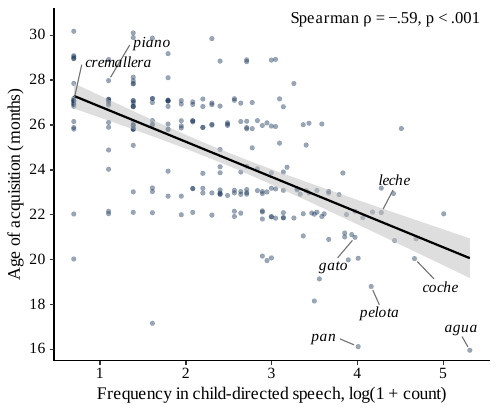}
\caption{Frequency in child-directed speech and age of acquisition for 225 Spanish nouns. Each point is a noun; points are jittered vertically by up to 0.2 months to reduce overlap. The line shows a linear fit with its 95\% confidence band. A few early- and late-learned nouns are labeled for illustration. The Spearman correlation is computed on raw counts.}
\label{fig:freq}
\end{figure}

Frequency was strongly related to AoA: nouns heard more often were learned earlier ($\rho = -.59$, $p < .001$; Figure~\ref{fig:freq}). Contextual diversity behaved almost identically ($\rho = -.59$, $p < .001$), which is unsurprising given that the two measures were very highly correlated ($\rho = .93$). When each was controlled for the other, frequency kept a small unique association with AoA ($\rho = -.16$, $p = .019$), whereas contextual diversity did not ($\rho = -.12$, $p = .084$). In the baseline regression, log frequency ($\beta = -.52$, $p < .001$) and word length ($\beta = .18$, $p = .002$) together explained 37.5\% of the variance in AoA. Longer words tended to be learned later.

\subsubsection{Surprisal}

\begin{figure*}[t]
\centering
\includegraphics[width=\textwidth]{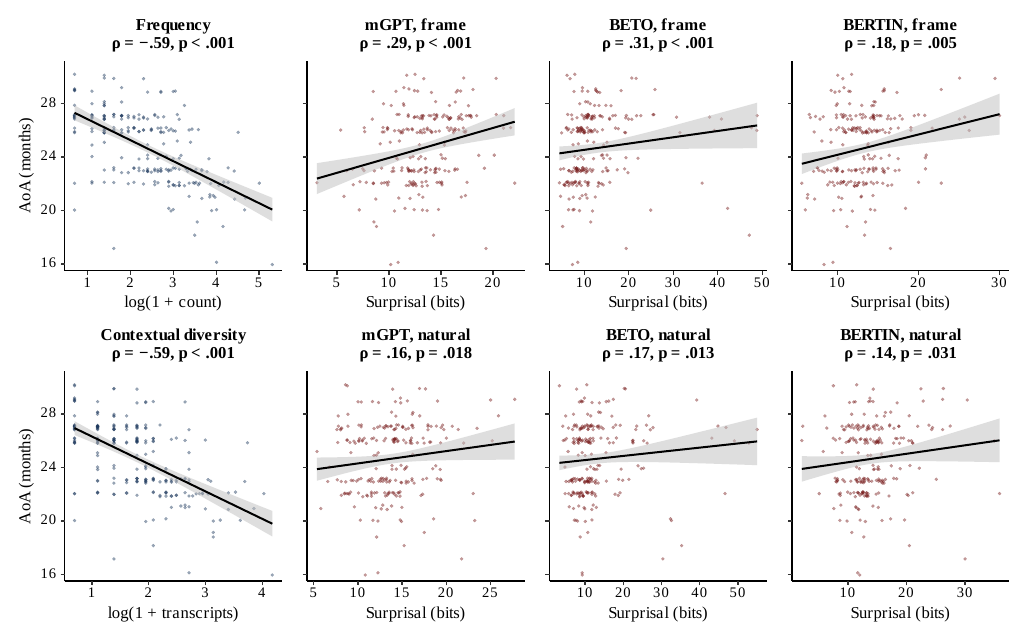}
\caption{All Study~1 predictors and age of acquisition. Top row: frequency and carrier-frame surprisal from the three models. Bottom row: contextual diversity and surprisal averaged over natural child-directed utterances. Each point is a noun (jittered vertically by up to 0.2 months). Lines show linear fits with 95\% confidence bands; correlations are Spearman's $\rho$ on untransformed values. Corpus measures are shown in blue, model surprisal in red.}
\label{fig:pred}
\end{figure*}

All six surprisal measures were positively correlated with AoA (Table~\ref{tab:corr}; Figure~\ref{fig:pred}): nouns that the models found less predictable tended to be learned later. The correlations were larger in the carrier frame ($\rho = .18$ to $.31$) than in natural contexts ($\rho = .14$ to $.17$). The three models agreed only moderately with one another (frame condition: $\rho = .49$ to $.53$), and frame and natural estimates for the same model agreed to different degrees (mGPT $\rho = .41$, BETO $.60$, BERTIN $.75$). Surprisal was also negatively correlated with frequency ($\rho = -.20$ to $-.38$): frequent nouns were more predictable for the models. The positive relation between surprisal and AoA could therefore reflect, at least in part, the fact that less predictable nouns are also less frequent.

\begin{table}[t]
\centering
\begin{threeparttable}
\caption{Study 1: Spearman correlations of each predictor with age of acquisition, and with log frequency.}
\label{tab:corr}
\footnotesize\setlength{\tabcolsep}{4pt}
\begin{tabular}{llrrrr}
\toprule
 & & \multicolumn{2}{c}{With AoA} & \multicolumn{2}{c}{With frequency} \\
\cmidrule(lr){3-4}\cmidrule(lr){5-6}
Predictor & Context & $\rho$ & $p$ & $\rho$ & \\
\midrule
Frequency & CDS corpus & $-.59$ & $<.001$ & -- & \\
Contextual div. & CDS corpus & $-.59$ & $<.001$ & $.93$ & \\
\addlinespace
mGPT surprisal & Frame & $.29$ & $<.001$ & $-.28$ & \\
BETO surprisal & Frame & $.31$ & $<.001$ & $-.38$ & \\
BERTIN surprisal & Frame & $.18$ & $.005$ & $-.30$ & \\
\addlinespace
mGPT surprisal & Natural & $.16$ & $.018$ & $-.20$ & \\
BETO surprisal & Natural & $.17$ & $.013$ & $-.24$ & \\
BERTIN surprisal & Natural & $.14$ & $.031$ & $-.24$ & \\
\bottomrule
\end{tabular}
\begin{tablenotes}\footnotesize
\item \textit{Note.} $N = 225$ (natural contexts: 224). Positive correlations with AoA mean that higher values go with later acquisition.
\end{tablenotes}
\end{threeparttable}
\end{table}

That is what the regressions showed. Once log frequency and length were in the model, no surprisal measure added more than 1.1\% of explained variance (Table~\ref{tab:incr}). Five of the six coefficients were far from significance. The sixth, BETO in natural contexts ($p = .049$), had the opposite sign to its bivariate correlation and did not survive correction for six tests (Holm $p = .29$).

\begin{table*}[t]
\centering
\begin{threeparttable}
\caption{Study 1: Variance in AoA explained by each surprisal measure beyond log frequency and word length.}
\label{tab:incr}
\small
\begin{tabular*}{\textwidth}{@{\extracolsep{\fill}}llcccccc}
\toprule
Model & Context & $R^2$ & $\Delta R^2$ & $\beta$ & $SE$ & $p$ & $p_{\text{Holm}}$ \\
\midrule
\multicolumn{2}{@{}l}{Frequency + length (baseline)} & .375 & -- & -- & -- & -- & -- \\
\addlinespace
mGPT & Frame & .379 & .004 & $.070$ & .056 & .214 & .804 \\
BETO & Frame & .381 & .006 & $-.083$ & .057 & .143 & .713 \\
BERTIN & Frame & .375 & .000 & $.003$ & .057 & .955 & .955 \\
mGPT & Natural & .376 & .002 & $-.041$ & .055 & .463 & .926 \\
BETO & Natural & .386 & .011 & $-.112$ & .056 & .049 & .292 \\
BERTIN & Natural & .379 & .005 & $-.071$ & .056 & .201 & .804 \\
\bottomrule
\end{tabular*}
\begin{tablenotes}\footnotesize
\item \textit{Note.} Ordinary least squares with all variables $z$-scored; $\beta$ is the standardized coefficient of surprisal. $N = 225$ for frames and 224 for natural contexts (baseline $R^2 = .375$ in both).
\end{tablenotes}
\end{threeparttable}
\end{table*}

\subsubsection{Robustness checks}

The rank-based partial correlations, controlling for log frequency and length, pointed the same way. Only mGPT in the carrier frame showed a small positive association with AoA ($\rho = .14$, 95\% CI [.01, .27], $p = .032$), which did not survive correction (Holm $p = .19$); the other five ranged from $-.03$ to $.09$ (all $p > .17$). Restricting the natural condition to the 154 nouns with at least five utterances did not change the picture (Table~\ref{tab:robust}). The bivariate correlations there were of similar size ($\rho = .14$ to $.17$), and none of the three measures explained significant variance beyond frequency and length, although BERTIN came close ($p = .055$, with a coefficient again opposite in sign to its bivariate correlation).

\begin{table}[t]
\centering
\begin{threeparttable}
\caption{Study 1: Natural-context surprisal for the 154 nouns with at least five utterances.}
\label{tab:robust}
\footnotesize\setlength{\tabcolsep}{4pt}
\begin{tabular}{lcccccc}
\toprule
Model & $\rho_{\text{AoA}}$ & $R^2_{\text{base}}$ & $R^2_{\text{full}}$ & $\Delta R^2$ & $\beta$ & $p$ \\
\midrule
mGPT & $.15$ & .346 & .349 & .004 & $-.063$ & .363 \\
BETO & $.17$ & .346 & .356 & .010 & $-.105$ & .134 \\
BERTIN & $.14$ & .346 & .362 & .016 & $-.135$ & .055 \\
\bottomrule
\end{tabular}
\begin{tablenotes}\footnotesize
\item \textit{Note.} Bivariate $p$ values: mGPT $.064$, BETO $.040$, BERTIN $.094$. Regressions as in Table~\ref{tab:incr}.
\end{tablenotes}
\end{threeparttable}
\end{table}

\subsection{Interim discussion}

Study~1 gives strong evidence that cumulative exposure is tied to when early words are learned. Nouns heard more often in child-directed speech were learned earlier, replicating a well-established result, and frequency kept a unique association with AoA when contextual diversity was controlled, while the reverse was not true.

The surprisal results are more nuanced than a simple null. Nouns that the models found less predictable were learned later, with all three models and in both kinds of context. But this relation was largely shared with frequency: none of the six measures explained reliable variance once frequency and length were known, and the few nominal effects were small, inconsistent in sign, and did not survive correction. The most defensible reading is that, in these data, contextual predictability carries little information about the timing of acquisition beyond cumulative exposure. This does not mean that children do not use expectations while learning. It means that a measure built to capture how predictable a word is on a given occasion adds little to explaining when it is learned, once we know how often the child has heard it.

This sets up Study~2. If the weak acquisition result simply reflected a failure of mGPT to capture anything real about Spanish, the same model should also do poorly with adult reading.

\section{Study 2: Surprisal and Adult Reading Times}

\subsection{Method}

\subsubsection{Participants and materials}

We used the Chilean Spanish subsample of the Multilingual Eye-movement Corpus, Wave~2 \citep[MECO;][]{siegelman2022,siegelman2025}, in which adult participants read short expository passages for comprehension while their eye movements were recorded. The subsample comprises 45 participants reading 12 passages. The dependent variable was total fixation duration, the sum of all fixations on a word, including first-pass reading and later re-reading; it was log-transformed to reduce skew. We excluded the first word of each passage, which has no preceding context, words that were missing from either frequency norm, and words that were skipped (no fixation). The analyzed data comprised 46,442 observations of 888 word types (median 981 observations per participant). Mean total fixation duration was 364 ms ($SD = 284$, median 281).

\subsubsection{Predictors}

Surprisal was computed with mGPT, the autoregressive model used in Study~1, without fine-tuning. For each word, the model was given the full preceding passage as context. Words were tokenized as in running text, each with its leading space, and word-level surprisal (in bits) was the sum over the word's subword tokens. To test the role of context length, we recomputed surprisal four more times with the context limited to the previous 20, 10, 5 or 3 words (across sentence boundaries). With full context, mean surprisal was 4.64 bits ($SD = 4.01$); with three words of context it was 7.48 bits ($SD = 4.91$).

To make sure that any surprisal effect was not tied to one frequency measure, we used two norms: EsPal \citep{duchon2013}, on the Zipf scale \citep{vanheuven2014}, and SUBTLEX-ESP \citep{cuetos2011}, a subtitle-based norm (log frequency). The two were highly correlated ($r = .95$), so rather than entering both in one model we fit parallel models with each. Word length was measured in letters.

\subsubsection{Statistical analysis}

We fit linear mixed-effects models of log total fixation duration with surprisal, frequency and word length as fixed effects and crossed random intercepts for participants and word types:
\begin{multline}
\log(\text{duration}) \sim \text{surprisal} + \text{frequency} + \text{length}\\
+ (1 \mid \text{participant}) + (1 \mid \text{word}).
\end{multline}
Model~A used EsPal frequency and Model~B SUBTLEX-ESP. Models were fit in R with lme4 \citep{bates2015} by restricted maximum likelihood, with $p$ values from Satterthwaite's approximation in lmerTest \citep{kuznetsova2017}. The surprisal effect was also tested with a likelihood-ratio test against the same model without surprisal (maximum likelihood fits). We did not include random slopes; with 45 participants this keeps the models stable, and all models converged without warnings. A third set of models, with all continuous variables $z$-scored, was used to compare the size of the three effects. To check for spillover, we added the surprisal of the previous word. Finally, the standardized model was refit with each of the four truncated-context surprisal estimates.

\subsection{Results}

\subsubsection{Surprisal, frequency and length}

With either frequency norm, surprisal was a reliable positive predictor of total fixation duration (Table~\ref{tab:meco}). Words that were less predictable from the preceding passage were fixated longer, over and above frequency, length, and the random variation between participants and words (likelihood-ratio tests: Model~A $\chi^2(1) = 167.7$, Model~B $\chi^2(1) = 175.0$; both $p < .001$). The estimate was nearly identical across the two norms. Higher frequency predicted shorter fixations and longer words predicted longer fixations, as expected.

\begin{table*}[t]
\centering
\begin{threeparttable}
\caption{Study 2: Mixed-effects models of log total fixation duration in MECO Wave 2 (Chilean Spanish).}
\label{tab:meco}
\small
\begin{tabular*}{\textwidth}{@{\extracolsep{\fill}}lcccc}
\toprule
 & \multicolumn{2}{c}{Model A (EsPal)} & \multicolumn{2}{c}{Model B (SUBTLEX-ESP)} \\
\cmidrule(lr){2-3}\cmidrule(lr){4-5}
Fixed effects & $b$ ($SE$) & $t$ & $b$ ($SE$) & $t$ \\
\midrule
Intercept & 5.451 (0.054) & 100.42 & 5.452 (0.047) & 115.63 \\
Surprisal (bits) & 0.0127 (0.0010) & 12.98 & 0.0128 (0.0010) & 13.27 \\
Frequency & $-0.0293$ (0.0070) & $-4.18$ & $-0.0359$ (0.0067) & $-5.34$ \\
Word length (letters) & 0.0547 (0.0025) & 22.13 & 0.0500 (0.0028) & 17.75 \\
\midrule
Random effects & Variance & $SD$ & Variance & $SD$ \\
\midrule
Word type & 0.0198 & 0.141 & 0.0195 & 0.140 \\
Participant & 0.0354 & 0.188 & 0.0354 & 0.188 \\
Residual & 0.2946 & 0.543 & 0.2946 & 0.543 \\
\bottomrule
\end{tabular*}
\begin{tablenotes}\footnotesize
\item \textit{Note.} 46,442 observations, 45 participants, 888 word types. Surprisal from mGPT with the full preceding passage. All fixed effects $p < .001$ (Satterthwaite).
\end{tablenotes}
\end{threeparttable}
\end{table*}

\begin{figure}[t]
\centering
\includegraphics[width=\columnwidth]{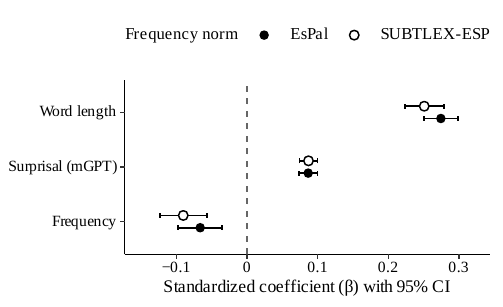}
\caption{Study 2: Standardized fixed effects on log total fixation duration, with 95\% confidence intervals, for the models with EsPal (filled) and SUBTLEX-ESP (open) frequency. Surprisal from mGPT with the full preceding passage.}
\label{fig:effects}
\end{figure}

In the standardized models (Figure~\ref{fig:effects}), word length had the largest effect (Model~A: $\beta = .27$, 95\% CI [.25, .30]), followed by surprisal ($\beta = .087$ [.073, .100]) and frequency ($\beta = -.066$ [$-.098$, $-.035$]). The same ordering held in Model~B (length $.25$, surprisal $.087$, frequency $-.090$).

The effect was not a spillover artifact. When the previous word's surprisal was added, the current word's surprisal remained significant ($b = 0.0102$, $SE = 0.0010$, $p < .001$), and the previous word's surprisal had a smaller effect of its own ($b = 0.0024$, $SE = 0.0008$, $p = .003$; $\chi^2(1) = 8.6$ with EsPal and 8.8 with SUBTLEX-ESP; $N = 46{,}122$).

\subsubsection{Context length}

\begin{figure}[t]
\centering
\includegraphics[width=\columnwidth]{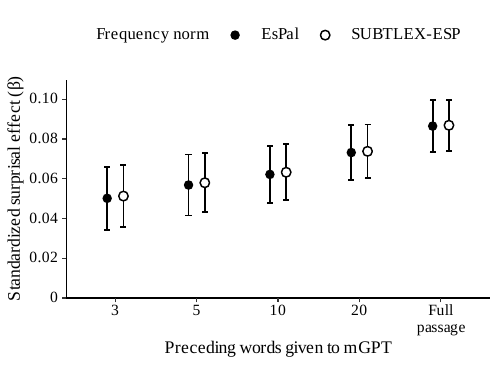}
\caption{Study 2: Standardized effect of mGPT surprisal on log total fixation duration as a function of how many preceding words the model was given, with 95\% confidence intervals, controlling for frequency and word length.}
\label{fig:context}
\end{figure}

Giving the model less context made the surprisal effect smaller, step by step, but never removed it (Table~\ref{tab:context}; Figure~\ref{fig:context}). With the previous 20 words, the standardized effect was .073; with 10 words, .062; with 5 words, .057; and with only the previous three words, .050 (95\% CI [.034, .066]; $\chi^2(1) = 37.8$, $p < .001$). The pattern was the same with SUBTLEX-ESP. Even with three words of context, slightly less than the four words that precede the noun in the Study~1 carrier frame, surprisal explained variance in reading times that frequency and length did not.

\begin{table*}[t]
\centering
\begin{threeparttable}
\caption{Study 2: Standardized surprisal effect as a function of the context given to mGPT.}
\label{tab:context}
\small
\begin{tabular*}{\textwidth}{@{\extracolsep{\fill}}lcccccc}
\toprule
 & \multicolumn{3}{c}{EsPal} & \multicolumn{3}{c}{SUBTLEX-ESP} \\
\cmidrule(lr){2-4}\cmidrule(lr){5-7}
Context & $\beta$ [95\% CI] & $\chi^2(1)$ & $\Delta$AIC & $\beta$ [95\% CI] & $\chi^2(1)$ & $\Delta$AIC \\
\midrule
Full passage & .087 [.073, .100] & 167.7 & 165.7 & .087 [.074, .100] & 175.0 & 173.0 \\
Last 20 words & .073 [.059, .087] & 106.7 & 104.7 & .074 [.060, .087] & 113.2 & 111.2 \\
Last 10 words & .062 [.048, .077] & 70.9 & 68.9 & .063 [.049, .077] & 76.9 & 74.9 \\
Last 5 words & .057 [.041, .072] & 51.9 & 49.9 & .058 [.043, .073] & 57.4 & 55.4 \\
Last 3 words & .050 [.034, .066] & 37.8 & 35.8 & .051 [.036, .067] & 41.9 & 39.9 \\
\bottomrule
\end{tabular*}
\begin{tablenotes}\footnotesize
\item \textit{Note.} Standardized mixed-effects models with frequency and word length; $\chi^2$ and $\Delta$AIC compare each model with the same model without surprisal (all $p < .001$). Positive $\Delta$AIC favors the model with surprisal.
\end{tablenotes}
\end{threeparttable}
\end{table*}

\subsection{Interim discussion}

Study~2 complements Study~1. Whereas surprisal added little to frequency and length in explaining AoA, mGPT surprisal predicted adult reading beyond the same predictors, with two frequency norms, after controlling for spillover, and with very short contexts. The context-length analysis matters for the comparison between studies: richer context strengthens the reading effect, as one would expect, but the effect does not depend on it. A difference in context length alone therefore cannot explain why surprisal predicts reading and not acquisition.

\section{General Discussion}

We asked whether surprisal from language models plays the same explanatory role at two points in the life of a word: when a child learns it, and when an adult reads it. The answer is no. In Study~1, frequency in child-directed speech was strongly related to AoA. Surprisal from three models was also related to AoA, but almost entirely through its overlap with frequency, and none of six measures explained reliable variance beyond frequency and word length. In Study~2, mGPT surprisal predicted adult fixation durations beyond frequency and length, with either frequency norm, after controlling for spillover, and even when the model saw only three words of context.

\subsection{Comparing the two studies}

\begin{figure}[t]
\centering
\includegraphics[width=\columnwidth]{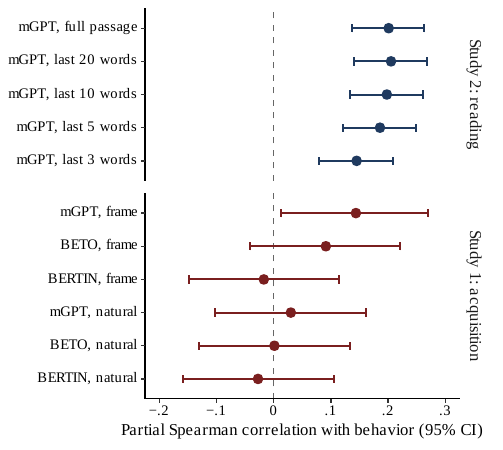}
\caption{The association between surprisal and behavior that is not explained by frequency and word length, in each study: Spearman partial correlations with 95\% confidence intervals. Study~2: mean log total fixation duration per word type (888 types; frequency from EsPal), mGPT with different context lengths. Study~1: AoA (225 nouns), three models in carrier frames and natural child-directed contexts. Positive values mean that more surprising words are read more slowly or learned later.}
\label{fig:unique}
\end{figure}

A difference between a significant and a non-significant result is not, by itself, evidence that two effects differ \citep{gelman2006}. To put the two studies on a common footing, we computed the same quantity in both: the rank-based partial correlation between surprisal and behavior, controlling for frequency and length, at the level of word types (Figure~\ref{fig:unique}). In both studies a positive value means that more surprising words are harder, that is, read more slowly or learned later. In reading, this association was clearly above zero at every context length ($\rho = .20$, 95\% CI [.14, .26] with the full passage; $.15$ [.08, .21] with three words). In acquisition, five of the six estimates were close to zero, and the sixth (mGPT in the carrier frame, $\rho = .14$) did not survive correction.

We do not claim that the two effects differ by a formal test, and it is worth being clear about why. The two studies use different populations, materials and dependent measures, so a direct comparison of correlations is only a rough guide. When we did compare the partial correlations directly (Fisher's $z$), the result depended on which measures were compared. The reading effect did not differ reliably from carrier-frame mGPT in acquisition ($z = 0.78$, $p = .43$), but it did differ from natural-context mGPT ($z = 2.30$, $p = .021$), and that difference fell short of significance when the reading context was limited to three words ($z = 1.54$, $p = .12$). What the data support with confidence is therefore more modest than a formal dissociation: surprisal has a reliable effect of its own on reading, and in acquisition we found no reliable effect of its own.

\subsection{Entrenchment and expectation}

The contrast fits a distinction between two properties of linguistic experience that are often treated as interchangeable. Frequency is an aggregate property, how often a child encounters a word across the whole input; surprisal is contextual, how strongly the preceding material predicts a particular occurrence. The results suggest that this distinction matters differently across development. Frequency was strongly tied to AoA and kept a unique association when contextual diversity was controlled, which converges with usage-based and entrenchment-based accounts of lexical development \citep{bybee2010,diessel2007,langacker1987,tomasello2003}. On these accounts, the property that decides when a word becomes established is the amount of experience with it, not the predictability of any single encounter.

Expectation-based theories of processing \citep{hale2001,levy2008} describe comprehenders who already have a lexicon and grammar and use context to anticipate what comes next. Study~2 fits that picture. Adult readers know the words, and when a word goes against their expectations they spend longer on it. Taken together, the two studies do not show that children fail to use prediction while learning. They show that a measure built to describe incremental prediction in a mature system adds little to explaining when words are learned, once we know how often children have heard them.

This is an interpretation of the pattern, not a demonstrated mechanism. Neither study manipulated frequency or predictability, and the account was formulated in light of the combined results.

\subsection{Error-driven learning}

For error-driven accounts \citep{ramscar2007,ramscar2010,ramscar2013}, the picture is less simple. On their simplest reading, these accounts predict that words met in less predictable contexts should be learned faster, leaving a negative relation between surprisal and AoA once frequency is controlled. We found no reliable relation in either direction: the bivariate association was positive, as if less predictable words were learned later, and it disappeared once frequency was controlled. This counts against a strong version of the error-driven view in which predictability shapes word learning independently of exposure. Two qualifications apply. First, frequency and predictability are correlated in children's input, so part of what frequency captures may itself reflect how predictable words are; our design cannot separate the two completely. Second, error-driven models define error over the specific cues and outcomes available to a learner in a given episode. Surprisal from models trained on adult text is a much coarser measure, and it may not track the expectations of a two-year-old. A cue-based measure of predictability might behave differently.

\subsection{Language models as models of human language}

These findings also bear on how language models are evaluated as models of human language. The same mGPT surprisal that predicts adult reading, even from three words of context, adds almost nothing to explaining when children learn nouns. This does not mean that mGPT is a poor model of Spanish, nor that its estimates are unrelated to development. It means that cognitive plausibility depends on the task and the population. A model trained mostly on adult text can capture what matters for mature comprehension without capturing what best explains how a vocabulary is built, and the input children receive is both smaller and differently structured. If the goal is to model acquisition, adult reading benchmarks are not enough; models should also be tested against developmental measures such as vocabulary composition, AoA and learning trajectories. More generally, it seems more useful to evaluate models against several populations and behavioral levels than to give them a single verdict of cognitive plausibility.

\subsection{Limitations}

Several limitations constrain these conclusions. First, both studies are correlational; they do not show that frequency causes earlier acquisition or that surprisal causes longer reading times. Second, the two datasets differ in population, register, variety and dependent measure: child-directed speech from four Spanish corpora and European Spanish CDI norms in Study~1, and expository texts read by Chilean adults in Study~2. These differences could contribute to the contrast between studies. Context length, at least, does not explain it, since the reading effect held with three words of context. Third, Study~1 used a single carrier frame, and for many nouns only a few natural contexts were available (median 9), so the surprisal estimates are noisy. Some masked-model estimates were also very large for words split into many subword pieces, which is one reason we report rank-based analyses alongside regressions. Fourth, AoA estimates from parental report are coarse, and the 225 nouns cover a fairly narrow range of ages. Finally, mGPT was the only model used in both studies; BETO and BERTIN converge with it for acquisition but were not evaluated on reading.

\subsection{Future directions}

Three steps would strengthen the comparison. The first is to estimate surprisal for acquisition from longer stretches of real child-directed speech, with the same model and estimator used for reading. The second is to test models trained on developmentally realistic input, such as those from the BabyLM challenge \citep{warstadt2023}, against the same AoA and reading benchmarks; this would show whether the weak acquisition result reflects the use of adult-trained models or a more basic difference between learning from exposure and processing by prediction. The third is to combine developmental and processing measures in the same variety and, ideally, in the same children followed over time, which would let us ask directly whether the information that predicts early learning gradually becomes the information that predicts skilled processing.

\subsection{Conclusion}

How often Spanish-learning children hear a noun strongly predicts when they learn it. How predictable the noun is for a language model is also related to when it is learned, but mostly because predictable nouns are also frequent; beyond frequency, surprisal explained no reliable variance. The same kind of measure predicts adult reading times, and does so even with very little context. Our reading of these results is that cumulative exposure drives the timing of early word learning, while contextual prediction shapes how words are processed once they are known. They also caution against treating a language model's cognitive plausibility as a single property, and point to the value of testing models against more than one population and behavioral benchmark.

\section*{Statements and Declarations}

\paragraph{Ethics.} This research used only publicly available, de-identified secondary data (Wordbank CDI norms, CHILDES transcripts, and the MECO Wave~2 eye-tracking corpus) and involved no new data collection. No institutional review was therefore required; the original data were collected and ethically approved by the respective corpus teams.

\paragraph{Funding.} This research received no specific grant from any funding agency in the public, commercial, or not-for-profit sectors.

\paragraph{Conflict of interest.} The author declares no conflict of interest.

\paragraph{Author contributions.} F.P.L. conceived the study, conducted the analyses, and wrote the manuscript.

\paragraph{Open science.} Both studies are exploratory analyses of secondary data and were not preregistered. The results should be considered provisional pending independent replication.

\paragraph{Data and code availability.} CDI norms are available from Wordbank (\url{https://wordbank.stanford.edu}), CHILDES corpora from TalkBank (\url{https://childes.talkbank.org}), and MECO Wave~2 (release 2.0) from the Open Science Framework (\url{https://osf.io/3527a/}). All derived data, the code that computes every surprisal measure, and the code that produces every number, table and figure in this paper are available at \url{https://github.com/fportillolo/frequency_surprisal_spanish}, together with a record of the corrections to the first version (\texttt{CORRECTIONS.md}).

\bibliographystyle{apacite}
\bibliography{references}

\end{document}